\documentclass[dvipsnames,format=sigconf,anonymous=false,review=false]{acmart}
\AtBeginDocument{%
  \providecommand\BibTeX{{%
    Bib\TeX}}}

\copyrightyear{2025}
\acmYear{2025}
\setcopyright{acmlicensed}\acmConference[GECCO '25]{Genetic and Evolutionary Computation Conference}{July 14--18, 2025}{Malaga, Spain}
\acmBooktitle{Genetic and Evolutionary Computation Conference (GECCO '25), July 14--18, 2025, Malaga, Spain}
\acmDOI{10.1145/3712256.3726359}
\acmISBN{979-8-4007-1465-8/2025/07}

\usepackage{algorithmic, algorithm}
\usepackage{graphicx}
\usepackage{multirow} 
\usepackage{subfigure}
\begin{document}


\title{SiamNAS: Siamese Surrogate Model for Dominance Relation Prediction in Multi-objective Neural Architecture Search}

\author{Yuyang Zhou}
\authornote{Both authors contributed equally to this research.}
\email{yuyang.zhou@nottingham.edu.cn}
\orcid{0009-0003-6726-0843}
\affiliation{%
  \institution{University of Nottingham Ningbo China}
  \city{Ningbo}
  \state{Zhejiang}
  \country{China}
}

\author{Ferrante Neri}
\authornotemark[1]
\email{f.neri@surrey.ac.uk}
\orcid{0000-0002-6100-6532}
\affiliation{%
  \institution{University of Surrey}
  \city{Guildford}
  \country{United Kingdom}}

\author{Yew-Soon Ong}
\email{asysong@ntu.edu.sg}
\orcid{0000-0002-4480-169X}
\affiliation{%
 \institution{Nanyang Technological University}
 \institution{Centre for Frontier AI Research, Institute of High Performance Computing, Agency for Science, Technology and Research}
 \city{Singapore}
 \country{Singapore}}


\author{Ruibin Bai}
\authornote{Corresponding author.}
\email{ruibin.bai@nottingham.edu.cn}
\orcid{0000-0003-1722-568X}
\affiliation{%
  \institution{University of Nottingham Ningbo China}
  \city{Ningbo}
  \state{Zhejiang}
  \country{China}
}

\renewcommand{\shortauthors}{Zhou et al.}

\begin{abstract}
Modern neural architecture search (NAS) is inherently multi-objective, balancing trade-offs such as accuracy, parameter count, and computational cost. This complexity makes NAS computationally expensive and nearly impossible to solve without efficient approximations. To address this, we propose a novel surrogate modelling approach that leverages an ensemble of Siamese network blocks to predict dominance relationships between candidate architectures. Lightweight and easy to train, the surrogate achieves 92\% accuracy and replaces the crowding distance calculation in the survivor selection strategy with a heuristic rule based on model size. Integrated into a framework termed SiamNAS, this design eliminates costly evaluations during the search process. Experiments on NAS-Bench-201 demonstrate the framework’s ability to identify Pareto-optimal solutions with significantly reduced computational costs. The proposed SiamNAS identified a final non-dominated set containing the best architecture in NAS-Bench-201 for CIFAR-10 and the second-best for ImageNet, in terms of test error rate, within 0.01 GPU days. This proof-of-concept study highlights the potential of the proposed Siamese network surrogate model to generalise to multi-tasking optimisation, enabling simultaneous optimisation across tasks. Additionally, it offers opportunities to extend the approach for generating Sets of Pareto Sets (SOS), providing diverse Pareto-optimal solutions for heterogeneous task settings.

\end{abstract}

\ccsdesc[500]{Computing methodologies~Discrete space search}
\keywords{Neural Architecture Search, Surrogate Assisted Models, Siamese Networks, Multi-objective Optimisation}


\maketitle

\section{Introduction}

Neural Architecture Search (NAS) automates the design and optimisation of neural networks, addressing the complex and resource-intensive task of model selection \cite{zoph2017nas}. Traditionally reliant on human expertise and trial-and-error, this process is often slow and yields suboptimal results \cite{liu2019darts, real2020evolutionary}. 
This automation not only accelerates development but also enables the discovery of models that outperform manually designed counterparts, driving progress in domains such as computer vision and natural language processing.

Modern NAS frameworks extend beyond optimising model accuracy, recognising that practical applications require balancing multiple objectives. Key considerations include minimising parameter count, computational complexity, inference latency, and energy consumption \cite{Shariatzadeh2023ASO}. These factors are particularly important for resource-constrained environments, such as smartphones and embedded systems, where efficiency is as critical as accuracy \cite{10004638}. Multi-objective NAS frameworks address these challenges by employing advanced optimisation methods, such as Pareto-based strategies and evolutionary algorithms, to discover architectures that strike an optimal balance among competing requirements. This holistic approach ensures that the resulting models are not only high-performing but also efficient and scalable for diverse deployment scenarios.

NAS poses distinct challenges for generative models, requiring multi-objective optimisation beyond traditional approaches. As shown in \cite{10336909}, evaluating generative adversarial networks (GANs) involves not only minimising complexity but also optimising metrics such as Fréchet Inception Distance (FID) and Inception Score (IS). This enables the design of efficient models that produce high-quality outputs.

NAS applications now extend well beyond image classification, covering areas such as medical diagnostics \cite{CHAIYARIN2024101565}, driver assistance \cite{9722706}, and natural language processing \cite{9762315}. These domains often involve highly complex and sometimes real-time constraints. As a result, enhancing computational efficiency through performance predictors—e.g., for accuracy, latency, and memory usage—has become crucial. At the same time, unified evaluation platforms like BenchENAS \cite{9697075} are increasingly vital for algorithm development. Moreover, the growing demand for flexible architectures capable of addressing multiple tasks via techniques like transfer learning further compounds the challenge \cite{9737315,10465622}.




All recent research directions of NAS as well as the challenges associated with them are characterised by their significant computational expense \cite{Real2017,Zoph2018,Real2019}. The inclusion of multiple objectives further exacerbates this computational cost. Specifically, multi-objective NAS requires the evaluation of a large number of solutions, which leads to the necessity of maintaining a large population in Evolutionary Multi-objective Optimisation (EMO). This added complexity makes it increasingly challenging to efficiently search for optimal architectures, as the number of function evaluations grows substantially with each additional objective \cite{10004638}. This holds true for various types of neural architectures, as emphasised in \cite{Booysen2024}.

For these reasons, a significant amount of effort in NAS research is dedicated to searching for new architectures while keeping the computational cost low, as reported in \cite{LIU2022100002}. The strategies for reducing computational cost can be classified as follows:

\begin{itemize} \item \textbf{Proxy-based NAS:} The computational cost is reduced by using shortcuts such as training on smaller datasets, training for fewer epochs, sharing the weights of a supernet, or evaluating scaled-down versions of architectures (e.g., \cite{9786036,liu2019darts}). \item \textbf{Surrogate-assisted NAS:} The computational cost is reduced by using an auxiliary model, often a machine learning model, that predicts the performance of candidate architectures based on historical evaluation data. Rather than directly evaluating each architecture, these methods train a surrogate model to estimate performance, enabling rapid evaluation and prioritisation of promising candidates (e.g., \cite{10.1007/978-3-030-58526-6_39,10.1145/3512290.3528703}). 
\end{itemize}

Within the surrogate-assisted category, a prominent approach does not attempt to reconstruct the original expensive objective function directly. Instead, it transforms the objective into a task that aims to preserve accurate rankings among candidate architectures. For instance, the study in \cite{Kyriakides2022} employs a Graph Convolutional Network to predict the relative ranking of architectures within an evolutionary cycle. Similarly, the works in \cite{9336721,9577556} develop models to predict pairwise rankings, facilitating the selection process within NAS. Following the same logic, the study in \cite{10.1007/978-981-97-5581-3_16} proposes leveraging isomorphism to inexpensively generate additional training data for graph-based encoding. Furthermore, an isomorphic score is introduced to predict the performance of candidate architectures.

The use of surrogate models in multi-objective NAS is inherently more complex than in single-objective cases, as it involves managing a greater number of interdependent components. These components can be computationally expensive, necessitating the use of strategies to mitigate these costs. The study in \cite{10.1007/978-3-030-58452-8_3} formulates NAS as a bi-level optimisation problem, employing surrogate models at both levels. At the upper level (architecture optimisation), four surrogate models are trained iteratively, and an adaptive mechanism selects the most accurate model for performance prediction. For the lower level (weights optimisation), a pre-trained supernet provides warm-start weights to candidate architectures, reducing the time required for evaluation. For real-time semantic segmentation, NAS is treated as a multi-objective problem in \cite{9916102}, using online and offline surrogate models to predict accuracy and latency, with a hierarchical prescreening strategy to balance both. \cite{10263998} employs cross-task transfer learning with iterative source selection to avoid negative transfer. \cite{10250852} proposes an online classifier for dominance prediction, supported by adaptive clustering and $\alpha$-domination to address class imbalance.

This paper, inspired by the ranking logic in single-objective NAS proposed in \cite{10.1007/978-981-97-5581-3_16} and the dominance predictor introduced in \cite{10263998}, presents a novel surrogate model designed to predict dominance between pairs of architectures within multi-objective NAS frameworks. The proposed model is a modified version of a Siamese network, utilizing its few-shot learning capability. This approach leverages the fact that the surrogate can be efficiently trained with a limited number of samples. 

The proposed surrogate model does not aim to approximate any objective function. Instead, it takes two input architectures and predicts the dominance relationship between them. This approach is proposed here because an accurate estimation of the dominance relationship between pairs of solutions is what is truly needed to map a population into non-dominated sets, making the approximation of an objective function unnecessary. Furthermore, a comparator-like surrogate model requires simpler architecture than an approximator and can be trained with a modest amount of data.

Similar to a Siamese network, the model consists of two identical sub-networks—here implemented as a simple Multi-Layer Perceptron (MLP) network. However, unlike a traditional Siamese network, the proposed model has three possible outcomes: dominance of the first architecture, dominance of the second architecture, or non-dominance between the two architectures. The surrogate can be efficiently trained using a few samples from the search space and seamlessly integrated into the non-dominated sorting mechanisms of many popular multi-objective optimisers.

The key contributions of this paper are as follows:
\begin{itemize}
\item A \textbf{novel surrogate model}, inspired by Siamese networks and leveraging \textbf{few-shot learning}, by training the surrogate model by a few ground truth performance metrics, is introduced for predicting the \textbf{dominance relationship} between pairs of architectures in multi-objective NAS.
\item The proposed surrogate model extends the standard Siamese network architecture by incorporating \textbf{three potential outputs} instead of two. These outputs represent the \textbf{three possible outcomes} of a dominance comparison: dominance of the first architecture, dominance of the second architecture, or non-dominance. This dominance prediction is made by \textbf{combining the outputs} of multiple Siamese surrogate blocks in an \textbf{ensemble model}, with the final prediction determined by two subsequent majority votes.
\item A novel \textbf{multi-objective NAS framework} based on the \textbf{Non-dominated Sorting Genetic Algorithm II (NSGA-II)} leveraging this surrogate model inspired by Siamese networks is proposed. The framework, namely SiamNAS, optimises \textbf{three objectives}: model accuracy, the number of parameters, and the number of floating-point operations (FLOPs) required to compute the output from a given input. This framework incorporates a modified survivor selection mechanism designed to enable efficient selection with minimal computational cost.

\end{itemize}

The remainder of this paper is structured as follows. Section \ref{sec:Problem} introduces the notation, problem formulation, and a brief overview of Siamese networks. Section \ref{sec:siamesesurrogatemodel} describes the architecture and components of the proposed surrogate model. Section \ref{sec:SiamNASframework} presents the SiamNAS framework and its integration with the surrogate. Section \ref{sec:expres} reports experimental results, and Section \ref{sec:conc} concludes the paper.




\section{Problem Formulation and Background}\label{sec:Problem}
If we denote a vector containing the architectural parameters as $\mathbf{x}$, identifying an architecture, $\Omega$ the set containing all the possible architectures, and $\mathbf{w}$ the vector of weights associated with a candidate architecture $\mathbf{x}$, NAS can be formulated as the following multi-objective optimisation problem \cite{10004638}.
\begin{equation}\label{eq:NAS}
\min_{\mathbf{x}}\left(\mathbf{f}_{e}\left(\mathbf{x};\mathbf{w^*}\right), \mathbf{f}_c\left(\mathbf{x}\right), \mathbf{f}_H\left(\mathbf{x}\right)\right)
\end{equation}
\begin{equation*}
    \text{s.t. }\mathbf{w^*} \in \text{argmin } \mathcal{L}_{trn}\left(\mathbf{x},\mathbf{w}\right)\text{,   } \mathbf{x}\in\Omega
\end{equation*}
\noindent where $\mathcal{L}_{trn}$ is the loss function used to train the architecture, $\mathbf{w^*}$ indicates the weights of the candidate architecture $\mathbf{x}$ after training, $\mathbf{f}_{e}$ is the vector of functions associated with the performance of the network, e.g. through its prediction error, $\mathbf{f}_c$ is the vector of functions associated with the complexity of the network (or more generally the model), e.g. the energy consumption of the model, $\mathbf{f}_H$ is the data structure associated with hardware performance.  

In this article, we perform the search of the architecture by optimising three objective functions, one of type $\mathbf{f}_{e}$ and two of type $\mathbf{f}_{c}$:
\begin{equation}\label{eq:fe}
    \mathbf{f}_{e}\left(\mathbf{x}\right)= 1-acc( \mathbf{x}, D)
\end{equation}
\begin{equation}\label{eq:fc}
    \mathbf{f}_{c}\left(\mathbf{x}\right)=\left( \#params(\mathbf{x}), FLOPS(\mathbf{x}) \right)
\end{equation}
\noindent where, $acc(\mathbf{x}, D)$ denotes the accuracy of the architecture $\mathbf{x}$ trained on the target dataset $D$, $\#params(\mathbf{x})$ represents the number of parameters in $\mathbf{x}$, and $FLOPs(\mathbf{x})$ refers to the number of floating-point operations required by $\mathbf{x}$ to calculate an output for a given input. It is noteworthy that, as documented in \cite{10250852}, $\#params(\mathbf{x})$ and $FLOPs(\mathbf{x})$ constitute conflicting objectives.

\subsection{Siamese Networks}
Siamese neural networks are a class of neural architectures designed to learn meaningful representations by comparing pairs of inputs. Introduced by Bromley et al. (1994) for signature verification tasks, these networks consist of two or more identical subnetworks that share weights and operate on different inputs in parallel. The shared architecture ensures that both inputs are transformed into comparable feature representations, making them particularly effective for tasks like similarity learning and metric-based decision-making. The output of a Siamese network is typically a similarity score, computed through a distance metric (e.g., Euclidean or cosine distance) applied to the learned representations. Variants of Siamese networks have since been adapted for various domains, including facial recognition \cite{chopra2005learning} and few-shot learning.


\section{Siamese Network Surrogate Model}\label{sec:siamesesurrogatemodel}



In this work, we propose a novel modification to the Siamese neural network architecture to serve as a surrogate model for predicting the dominance relationship in multi-objective problems between two candidate architectures. Conceptually, the proposed model consists of an ensemble of Siamese networks. Each Siamese network is composed of two identical multilayer perceptron (MLP) networks, which process input vectors representing a pair of architectures, denoted as \( \mathbf{x_1} \) and \( \mathbf{x_2} \). The proposed surrogate model evaluates \( \mathbf{x_1} \) and \( \mathbf{x_2} \) to determine which architecture, if any, dominates the other. Each Siamese network is trained separately and thus has its own set of weights. Within each Siamese network that composes the model, the weights of the two MLP networks are shared.

While Siamese networks were originally designed for tasks such as comparing facial features or fingerprints, we adapt this concept to the domain of dominance evaluation. The proposed surrogate model accommodates three possible outcomes: \( \mathbf{x_1} \preceq \mathbf{x_2} \), \( \mathbf{x_2} \preceq \mathbf{x_1} \), or non-dominated solutions.


\subsection{Encoding}

\begin{figure}[h]
    \centering
    \includegraphics[width=1\linewidth]{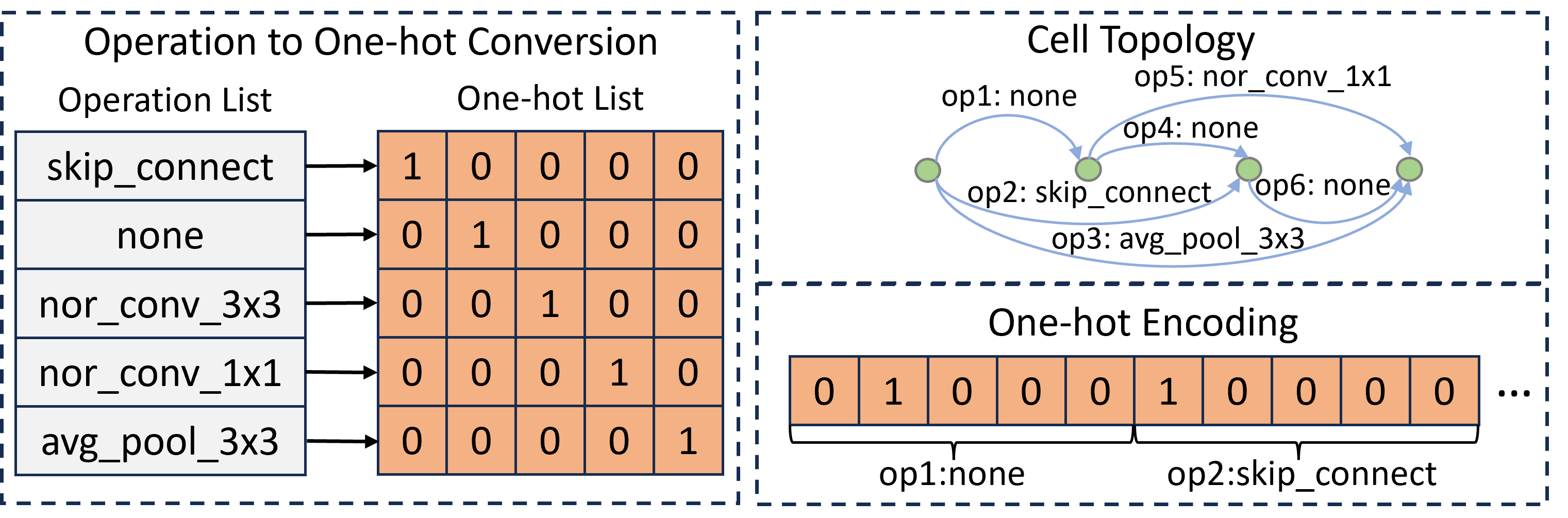}
    \Description{The left is an encoding list to map the original operation list of NAS-Bench-201 to a one-hot list. The right is an example of the one-hot encoding of an architecture in NAS-Bench-201.}
    \caption{Encoding a network from the NAS-Bench-201 search space as a binary vector \( \mathbf{x} \) to serve as input for the Siamese surrogate predictor.}
    \label{fig:encoding}
\end{figure}

As mentioned above, the input architectures are represented as encoded vectors, \( \mathbf{x_1} \) and \( \mathbf{x_2} \). To demonstrate the effectiveness of the proposed approach, this study utilizes the NAS-Bench-201 search space \cite{dong2020nasbench201}, where each candidate architecture is a convolutional neural network (CNN) block with six potential connections between four layers. Each connection can be assigned one of five possible operators, and the original representation as a 6-dimension vector specifies both the presence of connections and their associated operators. This representation is converted into a one-hot encoding format, consisting of a \( 6 \times 5 \) matrix. In this matrix, each row corresponds to a potential connection, and each column represents a specific operator. For each connection, a single bit in the row is set to 1, indicating the active operator, while all other bits are set to 0. Thus, each candidate architecture is represented as a binary vector of length $30$. An illustration of the encoding of a NAS-Bench-201 into a vector $\mathbf{x}$ is provided in Fig. \ref{fig:encoding}.

One-hot encoding is employed because it provides a clear, unbiased, and machine-readable format for categorical data \cite{naslib-2020}. This encoding avoids introducing any ordinal relationships between operators, ensuring that each operator is treated independently. Furthermore, it is compatible with the neural network architecture used in this study, facilitating effective processing and comparison of candidate architectures. By standardizing the input format, one-hot encoding ensures an efficient and semantically meaningful representation tailored to the task at hand.



\subsection{Multi-Layer Perceptron Networks within the Surrogate Model} \label{sec:siamese_surrogate}
As mentioned in the general description of the surrogate model, the proposed surrogate model consists of two identical MLP networks whose weights are shared between each other. Each MLP comprises one hidden layer, which is a fully connected layer with $32$ neurons, followed by the ReLU activation function. For each MLP, the binary input vectors \( \mathbf{x_1} \) and \( \mathbf{x_2} \) are processed independently. At the hidden layer, these inputs are transformed into real-valued embedding vectors, \( \mathbf{b_1} \) and \( \mathbf{b_2} \), each of length 32. These embeddings are then combined by computing their difference, resulting in the vector \( \mathbf{d} = \mathbf{b_1} - \mathbf{b_2} \). Naturally, in the case of identical architectures, i.e., \( \mathbf{x_1} = \mathbf{x_2} \), it follows that \( \mathbf{d} = \mathbf{o} \), where \( \mathbf{o} \) denotes the null vector.

The difference vector \( \mathbf{d} \) is processed by another MLP, which consists of two fully connected layers. The first layer comprises 32 neurons (processing  \( \mathbf{d} \)) followed by a ReLU activation function, while the second layer contains a single neuron that serves as the output. This output layer utilizes a sigmoid activation function, ensuring that the output is restricted to the range \((0, 1)\). 

The two MLP networks share identical weights and are trained using the Adam optimiser to minimizse the binary cross-entropy loss \cite{kingma2017adam}. The same optimiser and loss function are also employed to train the final part of the network, indicated as MLP Classifier in the yellow square depicted in Fig.~\ref{fig:implementationSiam}. 

\subsection{Ensemble-based Dominance Prediction} \label{sec:ensemble_ranking}

The output, as described in Section \ref{sec:siamese_surrogate}, is a scalar in range \(  (0, 1) \) which is rounded to $R \in \{0,1\}$. to represents the dominance relationship of \( \mathbf{x_1} \) and \( \mathbf{x_2} \). Specifically, \( R = 1 \) indicates that \( \mathbf{x_1} \preceq \mathbf{x_2} \), i.e. \( \mathbf{x_1} \) dominates \( \mathbf{x_2} \), while \( R = 0 \) implies that \( \mathbf{x_1} \preceq \mathbf{x_2} \) with probability 0, i.e., \( \mathbf{x_1} \not\preceq \mathbf{x_2} \). In the case of identical architectures, \( \mathbf{d} = \mathbf{o} \) and \( R = 0 \), which verifies the condition \( \mathbf{x_1} \not\preceq \mathbf{x_2} \). Finally, the rounded value \( R \) serves as the outcome of surrogate model prediction.

\begin{figure}
    \centering
    \includegraphics[width=1\linewidth]{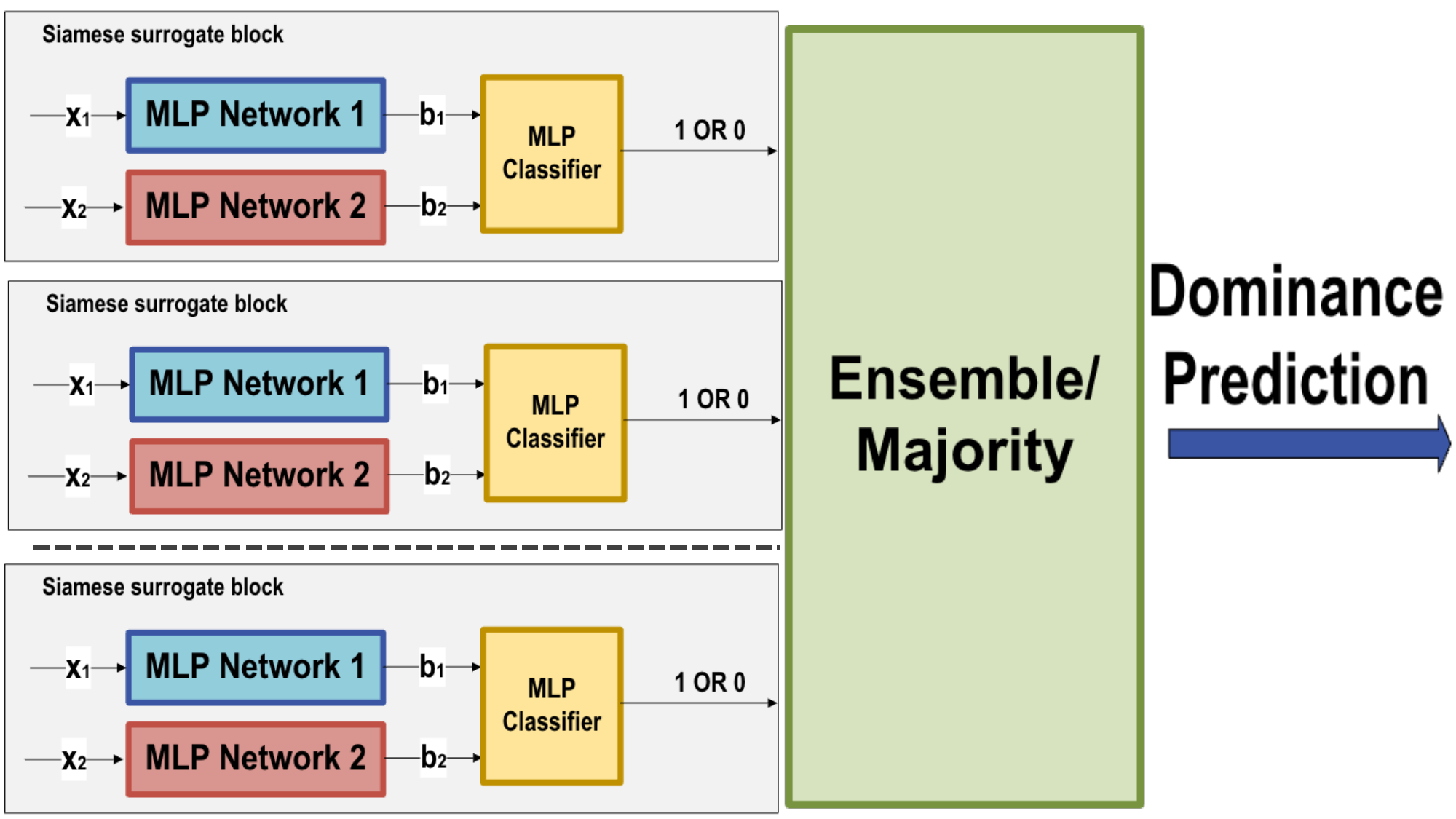}
    \Description{An illustration of how the dominance prediction of two input architectures are conducted. Inputs are fed into Siamese networks then classified by a MLP classifier. Several such rankers assembles an ensemble. The final dominance relationship is deducted by the majority output of the rankers in the ensemble. }
    \caption{Implementation of the Siamese network surrogate model: An ensemble of multiple Siamese network blocks is utilised to predict dominance relationships, with final predictions determined through a majority voting mechanism.}
    \label{fig:implementationSiam}
\end{figure}

In the proposed implementation, to enhance the performance of the model, multiple Siamese surrogate blocks, each composed of two identical MLP networks and a separate MLP classifier to process the embedding vector and calculate the output \( R \), are trained independently. These surrogate models are then combined in an ensemble approach to assess the dominance relationship between the two architectures. Specifically, \( \mathbf{x_1} \) and \( \mathbf{x_2} \) are fed into each surrogate model in the ensemble, generating predictions for \( R \). Let \( N_m \) denote the number of surrogate models, yielding predictions \( R_1, R_2, \ldots, R_{N_m} \). After rounding predictions to either 0 or 1, each surrogate model can be viewed as a classifier. If the majority of these classifiers predict an outcome of 1, then it is predicted that \( \mathbf{x_1} \preceq \mathbf{x_2} \). Otherwise, the order of the inputs (i.e., the candidate architectures $\mathbf{x_1}, \mathbf{x_2}$) is inverted, and the outputs are recalculated by the surrogate models to predict whether \( \mathbf{x_2} \preceq \mathbf{x_1} \). If the two consecutive predictions are \( \mathbf{x_1} \not\preceq \mathbf{x_2} \) followed by \( \mathbf{x_2} \not\preceq \mathbf{x_1} \), the model concludes that the two architectures do not dominate each other. An illustration of the implemented scheme is shown in Fig. \ref{fig:implementationSiam}.

\section{SiamNAS: A Multi-objective Algorithm for Neural Architecture Search}\label{sec:SiamNASframework}

The Siamese network surrogate model, described in Section \ref{sec:siamesesurrogatemodel}, is then integrated into the multi-objective NAS framework, here indicated as SiamNAS and presented in the following sections.


\subsection{Overall Framework}

\begin{algorithm}[ht]
\caption{SiamNAS Overall Framework} \label{alg:GA_framework}
\begin{algorithmic}
\STATE  \textbf{INPUT:} $\mathcal{S}$: search space , $N$: population size , $T$: number of generations, $R_c$: crossover rate, $R_m$: mutation rate, $M$: Siamese network  surrogate model, $N_s$: number of samples to train Siamese network base surrogate ensemble, $N_m$: number of Siamese surrogate models
\STATE \textbf{OUTPUT:} $\mathcal{P}$: A set of solutions on the same Pareto front
\end{algorithmic}
\begin{algorithmic}[1]
\STATE $D_s \leftarrow$ randomly sample $N_s$ architectures from the search space  $\mathcal{S}$ and calculate the three true objective function values for $\mathbf{f_e}$ $\mathbf{f_c}$ as per eq. (\ref{eq:fe}) and (\ref{eq:fc}) \# $Phase 1$
\STATE $M \leftarrow$ Use these $N_s$ architectures to train an ensemble of $N_m$ Siamese surrogate models, see Algorithm \ref{alg:train} 
\STATE $P \leftarrow$ initialize a population of size $N$ from $\mathcal{S}$ \# $Phase 2$
\STATE $t \leftarrow 0$
\WHILE{$t < T$ }
\STATE $P' \leftarrow$ binary tournament selection by $M$ from $P$
\STATE $P' \leftarrow$ apply uniform crossover to the architectures in $P'$ with crossover rate $R_c$
\STATE $P' \leftarrow$ apply flip mutation to each bit of the architectures in  $P'$ with mutation rate $R_m$
\STATE $P' \leftarrow$ check that all the architectures in $P'$ are meaningful
\STATE $P \leftarrow$ Biased-Selection($N$, $M$, $P$, $P'$) by Algorithm \ref{alg:selection}
\STATE $t \leftarrow t+1$
\ENDWHILE
\STATE $\mathcal{P} \leftarrow $ select the final population from $P$ and calculate the true objective function values \# $Phase 3$
\end{algorithmic}
\end{algorithm}

The pseudocode for the overall framework of the proposed SiamNAS is provided in Algorithm \ref{alg:GA_framework}. The framework is divided into three phases: Phase 1 (lines 1-2) focuses on building the ensemble of Siamese network surrogate models, see Section \ref{sec:training}; Phase 2 (lines 3-12) performs evolutionary NAS; and Phase 3 (line 13) obtains the final promising Pareto front from the last population.

It is important to note that in Phases 1 and 3, the true performance metrics of the candidate architectures are calculated, in this case queried from the NAS-Bench-201. In Phase 1, the number of architecture evaluations through true objective function calculations is equal to the number of candidate architectures selected for training, \( N_s \). In Phase 3, the number of queries corresponds to the size of the promising Pareto front in the last population, \( \mathcal{P} \), which is not greater than the population size \( N \). In total, since no true objective function calculations are performed in Phase 2, at most \( N_s + N \) architecture evaluations are made using true objective function calculations.

In Phase 2 (lines 3-12), evolutionary operations are performed on a population of \( N \) architectures within a multi-objective evolutionary framework. At each generation, solutions are selected for a mating pool using binary tournament selection, where pairwise comparisons are made with the Siamese surrogate model described in Section \ref{sec:siamesesurrogatemodel}. The selected solutions undergo uniform crossover and flip mutation with rates \( R_c \) and \( R_m \), respectively. A check is performed on the newly generated offspring architectures to ensure their validity and meaningfulness. In this case, the search space \( \mathcal{S} \) corresponds to that of the NAS-Bench-201 benchmark \cite{dong2020nasbench201}. Architectures that do not conform to the expected format are repaired by ensuring that each group of five bits, which is the one-hot representation of operations, contains exactly one bit set to 1, as illustrated in Fig. \ref{fig:encoding}. If multiple bits are set to 1 within a group, one is randomly chosen to remain 1, while the others are forced to 0. Conversely, if no bits are set to 1, a bit is randomly selected and flipped from 0 to 1. The survivor selection is then applied by means of a novel operator based on the proposed Siamese surrogate model. The details of this novel operator, namely Biased Selection, is outlined in  Section \ref{sec:siamese-selection}. 

\begin{figure*}[ht]
    \centering
    \includegraphics[width=\textwidth]{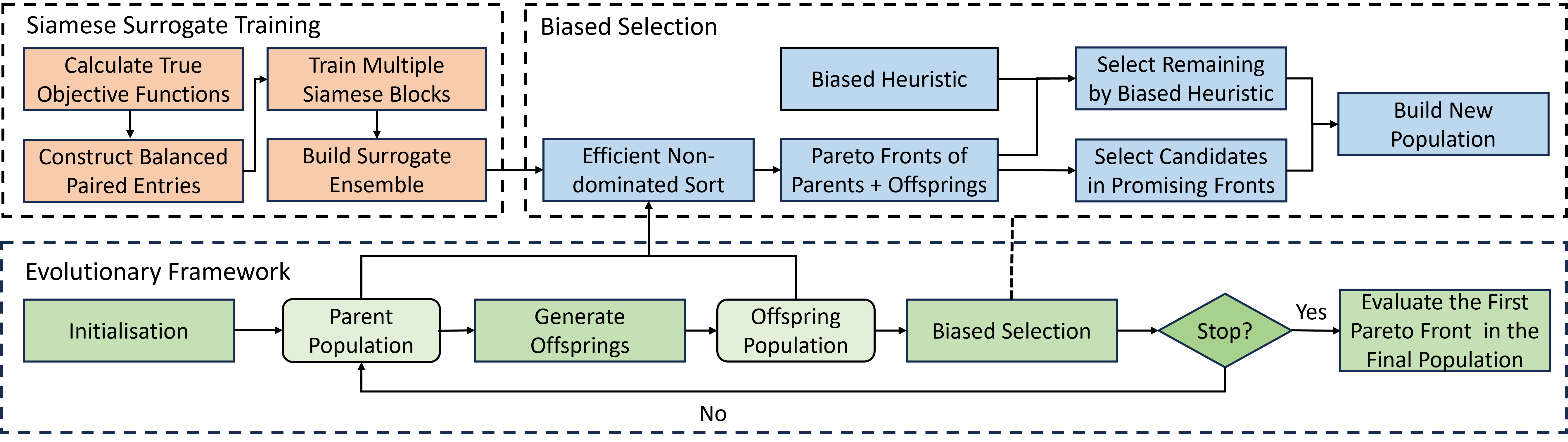}
    \Description{An overall framework showing how the SiamNAS works. In general it builds upon a multi-objective Genetic Algorithm framework, in which the dominance relationships are predicted by Siamese network based predictors instead of using the real data. It effectively reduce the cost on training neural networks.}
    \caption{Framework of SiamNAS: Green blocks represent the evolutionary operators, where offspring generation and biased selection leverage the surrogate model for dominance-based comparisons. Operations related to training the surrogate model are highlighted in orange, while the biased selection process is marked in blue.
}
    \label{fig:SiamNASFramework}
\end{figure*}

The scheme illustrating the functionality of the SiamNAS framework is presented in Fig. \ref{fig:SiamNASFramework}.

\subsection{Data Management and Training of the Siamese Network Surrogate Model}\label{sec:training}

As detailed in Phase 1 (lines 1--2) of Algorithm \ref{alg:GA_framework}, \( N_s \) candidate architectures are sampled from the search space, and their true objective function values are calculated. These architectures form a dataset to train the Siamese surrogate model. However, since the proposed surrogate model requires two inputs, pairs of inputs must be constructed. Consequently, from these \( N_s \) candidate architectures evaluated through the true objective functions, \( N_s \times N_s \) possible training entries can be generated. This approach significantly enlarges the available data for training without requiring additional evaluations of true objective functions.





With this dataset of \( N_s \times N_s \) entries, the Siamese surrogate model is trained by following Algorithm \ref{alg:train}. This algorithm is divided into three main stages: pair construction, pair reassignment, and training. In the first stage (lines 5--6), the \( N_s \) candidate architectures comprising \( D_s \) are shuffled to generate a shuffled dataset \( D_p \). The generic elements \( D_s^i \) and \( D_p^j \) are then paired to create a dataset of pairs, \( D'_t \), as shown in line 4.

During the creation of pairs, it is crucial to address the issue of class label imbalance, which is particularly important for surrogate models predicting dominance in multi-objective optimisation. As highlighted in \cite{10250852}, ensuring that positive class, the cases where  \( \mathbf{x_1} \preceq \mathbf{x_2} \) (scoring \( 1 \)), are roughly balanced with negative class, the cases where \( \mathbf{x_1} \not\preceq \mathbf{x_2} \) (scoring \( 0 \)), is essential for accurate training of the surrogate model. It is worth noting that, even with just three objectives, a simple random sampling of pairs is likely to result in a dataset dominated by negative class, exacerbating the class imbalance problem.
To mitigate the issue of class imbalance, the constructed pairs are reassigned to reduce the proportion of negative labels (label 0) to approximately 50\% (lines 7--14). A resampling rate, \( \theta \), is calculated according to the proportion of negative pairs in the dataset (lines 7--8). Each negative pair is then given a chance, with probability \( \theta \), to have its second candidate replaced. This replacement continues until a candidate is found that is dominated by the first candidate in the original pair (lines 9--14). This reassignment process ensures a more balanced dataset, which is essential for the effective training of the surrogate model.



After balancing the proportion of negative and positive pairs, labels \( l \) are assigned to all pairs. If the first candidate in a pair dominates the second candidate, a label of 1 is assigned; otherwise, a label of 0 is assigned. These pairs of candidates, \( D_t \), together with their corresponding labels, \( l \), are used to train one Siamese surrogate block \(m \in \{m_1, m_2, \ldots\} \) in the ensemble. Each Siamese surrogate block is trained by different pairs of candidates $D_t$, constructed by the same true objective function values $D_s$. Once trained, these blocks collectively form the Siamese Network Surrogate Model, \( M \).

\begin{algorithm}
\caption{Train-Surrogate-Ensemble} \label{alg:train}
\begin{algorithmic}
\STATE  \textbf{INPUT:} $M$: Siamese network base surrogate ensemble, $D_s$: samples from search space for training, $N_m$: number of Siamese surrogate models
\STATE \textbf{OUTPUT:} $M$: trained Siamese Network Surrogate Model (based on the ensemble of Siamese network blocks) 
\end{algorithmic}
\begin{algorithmic}[1]
\FOR{$r=1 \ldots N_m$}
\STATE $m_r \leftarrow$ initialize Siamese surrogate model 
\STATE $D_t \leftarrow \emptyset$
\FOR{$i=1\ldots 100$}
\STATE  $D_p \leftarrow$ randomly shuffle $D_s$
\STATE $D_t' = \{(D_s^1, D_p^1),\ldots,(D_s^{N_s}, D_p^{N_s})\} \leftarrow$ combine each candidate in $D_s$ and $D_p$ to build pairs of data
\STATE $\sigma \leftarrow $ proportion of pairs in $D_t'$ where $D_s^*$ dominates $D_p^*$
\STATE $\theta \leftarrow 0.5/(1-\sigma)$
\FOR{$j=1,\ldots,N_s$}
\STATE $r \in \mathbb{R} \leftarrow$ uniformly distributed random value in $[0,1]$
\IF{$D_s^j$ does not dominate $D_p^j$ and $r< \theta$}
\STATE $D_p^j \leftarrow$ find a $D_p^k \in D_s$ that $D_s^j$ dominates $D_p^k$ if exists
\ENDIF
\ENDFOR
\STATE $D_t \leftarrow D_t \cup D_t'$
\ENDFOR
\STATE $l \leftarrow$ assign label to each pair in $D_t$, where the label is $1$ if the first candidate in pair dominates the other, otherwise $0$
\STATE $m_r \leftarrow$ train Siamese surrogate model by $D_t$ and $l$
\ENDFOR 
\STATE $M \leftarrow$ build ensemble by trained Siamese surrogate models $m_1 \ldots m_R$
\end{algorithmic}
\end{algorithm}

\subsection{Biased Selection Criterion} \label{sec:siamese-selection}
With reference to line 10 of Algorithm \ref{alg:GA_framework}, a modified survivor selection strategy that leverages the trained surrogate model is proposed. First, the population of parents, \( P \), is merged with the offspring population, \( P' \), resulting in an auxiliary population, \( Q \). This population, \( Q \), is then divided into non-dominated sets (fronts) by applying the efficient non-dominated sorting method described in \cite{ENS}, where each comparison is performed using the proposed Siamese network surrogate model (lines 1-2 of Algorithm \ref{alg:selection}).

In the classical multi-objective optimisation algorithm NSGA-II \cite{NSGA-II}, the crowding distance is employed as a secondary criterion, following non-dominated sorting, to decide which candidates should be retained for the next generation. However, in our case, as the true objective function values are not available, the calculation of the crowding distance is infeasible without resorting to true objective function evaluations, which would negate the computational efficiency achieved thus far.


To address the absence of crowding distance in NSGA-II, we propose a simple heuristic method detailed in Algorithm \ref{alg:selection} that slightly amends the popular NSGA-II selection mechanism outlined in \cite{NSGA-II}. In this study, instead of retaining candidates with the largest crowding distance, we prioritise the selected candidates with the largest number of parameters during the survivor selection phase (lines 8–12). The rationale behind this choice is twofold:  
\begin{enumerate}
    \item The number of parameters is a training-free metric. Acquiring this information incurs significantly lower computational cost compared to evaluating the true objective functions such as accuracy.
    \item Empirical evidence suggests that larger neural network models, characterised by a greater number of parameters, often exhibit superior performance.
\end{enumerate}

In other words, the preference for larger models is employed as a proxy for performance, serving to counterbalance the minimisation of the number of parameters. Conversely, favouring smaller models within the non-dominated set would bias the search towards architectures likely to exhibit unsatisfactory performance.

\begin{algorithm}
\caption{Biased-Selection} \label{alg:selection}
\begin{algorithmic}
\STATE  \textbf{INPUT:} $N$: population size, $M$: Siamese network base surrogate ensemble, $P$: parent population, $P'$: offspring population
\STATE \textbf{OUTPUT:} $Q$: New population
\end{algorithmic}
\begin{algorithmic}[1]
\STATE $Q' \leftarrow P \cup P'$ 
\STATE $\mathcal{F} = \{F_1, F_2,...,F_k\} \leftarrow$ sequential efficient non-dominated sort $Q'$ by $M$
\STATE $Q \leftarrow \emptyset$, $i=1$
\WHILE{$|Q|+|F_i| \leq N$ and $i\leq k$}
\STATE $Q \leftarrow Q \cup F_i$
\STATE $i \leftarrow i+1$
\ENDWHILE
\IF{$|Q| < N$}
\STATE $N_r \leftarrow N - |Q|$ 
\STATE $F_i'=\{s_1, s_2,...\} \leftarrow $ sort $F_i$ in descending order by the number of trainable parameters of the candidate neural architecture
\STATE $Q \leftarrow Q \cup \{s_1,...s_{N_r}\}$
\ENDIF
\end{algorithmic}
\end{algorithm}




\section{Experimental Results}\label{sec:expres}

The proposed SiamNAS framework was evaluated on the NAS-Bench-201 \cite{dong2020nasbench201} search space using three datasets: CIFAR10 \cite{Krizhevsky2009cifar}, CIFAR100 \cite{Krizhevsky2009cifar}, and ImageNet-16-120 \cite{russakovsky2015imagenetlargescalevisual}. To train the Siamese network surrogate model, a sample of \( N_s = 600 \) architectures with true objective function values, namely the train accuracy, the number of parameters and the FLOPs, was selected. Training was performed using the ADAM optimiser \cite{kingma2017adam} over 20 epochs, with a learning rate of 0.001, a batch size of 100, and binary cross-entropy as the loss function. The ensemble consisted of seven Siamese network blocks.

Within the evolutionary framework, the population size \( N \) was set to 50, the crossover rate \( R_c \) to 0.7, the mutation rate to 0.1, and the total number of generations \( T \) to 2000. Each experiment was repeated 10 times to ensure robustness, with all computations performed on the same machine equipped with a single NVIDIA GeForce RTX 4090 GPU card.


\subsection{Ablation Studies}

The design of SiamNAS and its hyperparameter settings were determined through comprehensive ablation studies. In this section, we examine the impact of the ensemble size, $N_m$, on accuracy. The experimental settings of the ablation study follow the same as those specified in Section \ref{sec:expres}, while the number of Siamese blocks in one ensemble is altered from 1 to 13. The results, presented in Table \ref{tab:ensemble size}, indicate that using seven Siamese network blocks yields at least the second best results in terms of both accuracy and F1 score. Furthermore, increasing the ensemble size beyond this point does not result in any significant improvement.

\begin{table}[]
\centering
\caption{Ablation study of the impact of ensemble size $N_m$ on the prediction accuracy and F1 score on of the Siamese network surrogate model over NAS-Bench-201 and for CIFAR10, CIFAR100, and ImageNet-16-120.}\label{tab:ensemble size}
\resizebox{0.9\linewidth}{!}{%
\begin{tabular}{c|cc|cc|cc}
\hline
\multirow{2}{*}{$N_m$} & \multicolumn{2}{c|}{CIFAR10}  & \multicolumn{2}{c|}{CIFAR100} & \multicolumn{2}{c}{ImageNet-16-120} \\ \cline{2-7} 
                       & \multicolumn{1}{c|}{Accuracy} & F1 Score & \multicolumn{1}{c|}{Accuracy} & F1 Score & \multicolumn{1}{c|}{Accuracy}    & F1 Score    \\ \hline 
1                      & \multicolumn{1}{c|}{90.59}    & 0.70     & \multicolumn{1}{c|}{90.78}    & 0.70     & \multicolumn{1}{c|}{90.40}       & 0.73        \\
3                      & \multicolumn{1}{c|}{92.00}    & 0.74     & \multicolumn{1}{c|}{91.04}    & 0.71     & \multicolumn{1}{c|}{91.74}       & 0.76        \\
5                      & \multicolumn{1}{c|}{92.64}    & 0.76     & \multicolumn{1}{c|}{91.68}    & 0.72     & \multicolumn{1}{c|}{92.06}       & 0.77        \\
7                      & \multicolumn{1}{c|}{92.83}    & 0.76     & \multicolumn{1}{c|}{91.61}    & 0.72     & \multicolumn{1}{c|}{92.57}       & 0.79        \\
9                      & \multicolumn{1}{c|}{93.02}    & 0.77     & \multicolumn{1}{c|}{91.36}    & 0.71     & \multicolumn{1}{c|}{92.57}       & 0.79        \\
11                     & \multicolumn{1}{c|}{92.77}    & 0.76     & \multicolumn{1}{c|}{91.42}    & 0.72     & \multicolumn{1}{c|}{92.38}       & 0.78        \\
13                     & \multicolumn{1}{c|}{92.83}    & 0.76     & \multicolumn{1}{c|}{91.49}    & 0.72     & \multicolumn{1}{c|}{92.45}       & 0.78      
\\ \hline 
\end{tabular}
}
\end{table}

Furthermore, we present the impact of the training dataset size, \( N_s \), on the accuracy of the predictive Siamese network surrogate model. While all other experimental settings are the same as introduced at the beginning of Section \ref{sec:expres}, the surrogate model was tested with \( N_s \) ranging from 0 to 1000, and the results are summarised in Table \ref{tab:ablationNs}. The findings reveal a trade-off between the model accuracy/F1 and the number of architectures for which true objective function values are calculated. Additionally, the corresponding GPU computational costs are provided. Notably, for \( N_s = 600 \), the surrogate model achieves an accuracy of approximately \( 92\% \), with further increases in \( N_s \) offering no significant improvement in predictive accuracy.



\begin{table}[] 
\centering
\caption{Comparison of surrogate models prediction performance with different number of true objective function calculations, i.e., $N_s$ on NAS-Bench-201.}\label{tab:ablationNs}
\resizebox{1\linewidth}{!}{%
\begin{tabular}{c|c|cc|cc|cc}
\hline
\multirow{2}{*}{$N_s$}        & Cost          & \multicolumn{2}{c|}{CIFAR10} & \multicolumn{2}{c|}{CIFAR100} & \multicolumn{2}{c}{ImageNet-16-120} \\ \cline{3-8} 
 & (GPU Days) & \multicolumn{1}{c|}{Accuracy}     & F1 Score    & \multicolumn{1}{c|}{Accuracy} & F1 Score & \multicolumn{1}{c|}{Accuracy}        & F1 Score        \\ \hline
None             & N/A           & \multicolumn{1}{c|}{15.56}             &     0.24        & \multicolumn{1}{c|}{11.65}    & 0.21  & \multicolumn{1}{c|}{14.21}  & 0.25                \\
100              & 1.67e-3       & \multicolumn{1}{c|}{86.36}        & 0.48        & \multicolumn{1}{c|}{86.43}    & 0.53     & \multicolumn{1}{c|}{88.54}  & 0.62                \\
200              & 1.99e-3        & \multicolumn{1}{c|}{88.28}        & 0.60        & \multicolumn{1}{c|}{90.14}    & 0.62     & \multicolumn{1}{c|}{89.44}   &  0.62               \\
400              & 2.77e-3       & \multicolumn{1}{c|}{90.91}        & 0.71        & \multicolumn{1}{c|}{91.68}    & 0.69     & \multicolumn{1}{c|}{91.04}    &  0.71              \\
600              & 3.65e-3        & \multicolumn{1}{c|}{92.45}        & 0.71        & \multicolumn{1}{c|}{93.41}    & 0.77     & \multicolumn{1}{c|}{91.87}             & 0.73                \\
800              & 4.56e-3        & \multicolumn{1}{c|}{92.19}        & 0.75        & \multicolumn{1}{c|}{93.09}    & 0.75     & \multicolumn{1}{c|}{92.70}    & 0.80                \\
1000             & 5.63e-3        & \multicolumn{1}{c|}{92.70}             &    0.78         & \multicolumn{1}{c|}{92.57}    & 0.74 & \multicolumn{1}{c|}{92.57} & 0.77      
\\ \hline
\end{tabular}
}
\end{table}

\subsection{Results on NAS-Bench-201}

The performance of the proposed SiamNAS on NAS-Bench-201 was evaluated across CIFAR-10, CIFAR-100, and ImageNet-16-120 datasets and compared against twenty-six state-of-the-art algorithms from the literature. Among these competitors, eight algorithms search within the NAS-Bench-201 space without surrogate models (upper part of Table \ref{tab:compare_201}), employing alternative mechanisms such as weight inheritance to reduce computational cost. The remaining eighteen algorithms incorporate surrogate models in their search process (lower part of Table \ref{tab:compare_201}). The detailed comparison results are presented in Table \ref{tab:compare_201} where Oracle indicates the performance of the best architecture searchable in the NAS-Bench-201 space.


Experimental results on the accuracy of the searched models demonstrate that SiamNAS achieves excellent performance on CIFAR-10, successfully identifying the theoretical optimum and outperforming all competitors. Additionally, SiamNAS exhibits competitive performance on CIFAR-100 and ImageNet-16-120. To further validate its effectiveness, the Siamese network surrogate models trained on CIFAR-10 was applied to the classification tasks on the CIFAR-100 and ImageNet-16-120 datasets. The experimental results indicate evidence of positive transfer, with the architecture achieving very good performance on both datasets. Interestingly, the results obtained through transfer learning slightly surpass those achieved when using the datasets directly, where the ImageNet-16-120 error rate is the second best among all methods.

It is worthwhile to highlight several key points. First, SiamNAS is a multi-objective NAS algorithm, whereas most of the compared algorithms are single-objective. In this context, the performance of SiamNAS is even more impressive. Second, the number of architectures evaluated using true objective functions is 600 for SiamNAS, compared to as few as 100 for other surrogate-based models (and as low as 90 in the case of ReNAS~\cite{xu_renas_2021}). This discrepancy arises because the sorting process in multi-objective optimization is inherently more complex than in the single-objective case. Lastly, the algorithm runtime (excluding the time to perform the evaluations of architectures through true objective functions) of SiamNAS is significantly lower than that of all its competitors. For instance, SiamNAS has a runtime of roughly 0.01 GPU days, whereas BANANAS~\cite{white_bananas_2021} requires 2 GPU days as shown in Table \ref{tab:runtime_201}.

\begin{figure}[h]
    \centering
    \includegraphics[width=\linewidth]{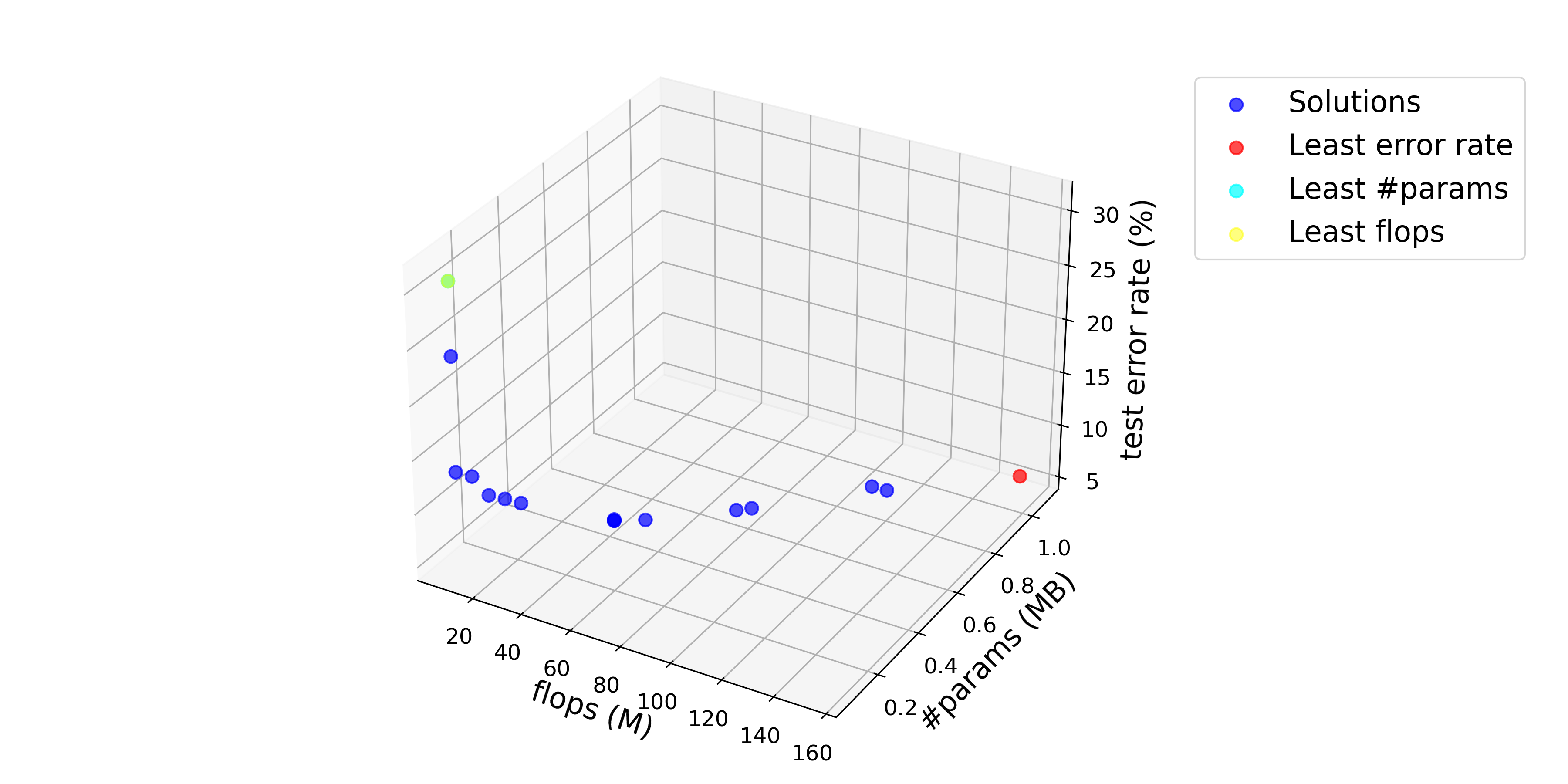}
    \Description{It shows a Pareto front found by SiamNAS.}
    \caption{The solutions in a Pareto front found by SiamNAS on CIFAR10 dataset on NAS-Bench-201.}.
    \label{fig:3dplot}
\end{figure}

An example of the final non-dominated set identified by SiamNAS for the CIFAR10 dataset is illustrated in Fig. \ref{fig:3dplot}. The architecture with the highest accuracy is observed to have a significantly large number of parameters (measured in megabytes) and an exceptionally high number of FLOPs. Specifically, this architecture is characterized by $\#\text{params} = 1.074$ MB, $\text{FLOPs} = 1.533 \times 10^2 $ M, and $1 - \text{acc} = 5.627 $. Conversely, the architectures with minimal FLOPs and parameter counts, which are clustered closely in this example, exhibit suboptimal performance. These architectures are defined by $\#\text{params} = 7.331 \times 10^{-2}$ MB, $\text{FLOPs} = 7.783$ M, and $1-\text{acc} = 3.078 \times 10^1$. A solution with a balanced trade-off among the three objectives is marked in blue and associated with the objective function values  $\#\text{params} = 3.444 \times 10^{-1}$ MB, $\text{FLOPs} = 4.711\times 10^1$ M, and $1-\text{acc} = 6.837$.

\begin{table}[]
\centering
\caption{Comparison of NAS algorithms on NAS-Bench-201.}
\label{tab:compare_201}
\resizebox{0.98\linewidth}{!}{%
\begin{scriptsize}
\begin{tabular}{c|c|c|c}
\hline
\multirow{2}{*}{NAS Algorithm}                             & CIFAR10                & CIFAR100                & ImageNet-16-120\\ \cline{2-4} 
                                                     & Test Error Rate        & Test Error Rate         & Test Error Rate         \\ \hline
RSPS~\cite{li_random_2020}                           & 8.95$\pm$0.66          & 31.74$\pm$0.96          & 59.31$\pm$0.36          \\
SETN~\cite{9009811}                                  & 7.28$\pm$0.73          & 30.64$\pm$1.72          & 60.49$\pm$0.33          \\
ENAS~\cite{pham_efficient_2018}                      & 6.24$\pm$0.00          & 29.33$\pm$0.62          & 58.56$\pm$0.00          \\
FairNAS~\cite{chu_fairnas_2021}                      & 6.77$\pm$0.18          & 29.00$\pm$1.46          & 57.81$\pm$0.31          \\
DARTS~\cite{chu2021darts}                            & 6.20$\pm$0.40          & 28.47$\pm$1.51          & 54.88$\pm$0.82          \\
BOHB~\cite{falkner_bohb_2018}                        & 6.06$\pm$0.28          & 28.00$\pm$0.86          & 54.30$\pm$0.86          \\
REINFORCE~\cite{zoph_neural_2017}                    & 6.10$\pm$0.26          & 28.14$\pm$0.89          & 54.36$\pm$0.78          \\
GDAS~\cite{8953848}                                  & 6.77$\pm$0.58          & 31.83$\pm$2.50          & 60.60$\pm$0.00          \\
\hline
NAR~\cite{guo_generalized_2022}                      & 5.67$\pm$0.05          & 27.11$\pm$0.37          & \textbf{53.34$\pm$0.23} \\
GenNAS-N~\cite{li_generic_2021}                      & 5.82$\pm$0.10          & 27.44$\pm$0.74          & 54.41$\pm$0.54          \\
NASWOT~\cite{mellor_neural_2021}                     & 7.04$\pm$0.81          & 30.02$\pm$1.22          & 55.56$\pm$2.10          \\
MbML-NAS (RF)~\cite{pereira_neural_2023}             & 6.64$\pm$0.20          & 29.67$\pm$0.85          & 57.01$\pm$4.21          \\
MbML-NAS (GB)~\cite{pereira_neural_2023}             & 6.97$\pm$0.52          & 29.98$\pm$1.17          & 55.72$\pm$1.42          \\
NASWOT~\cite{mellor_neural_2021}                     & 7.19$\pm$0.99          & 30.52$\pm$1.70          & 56.90$\pm$3.16          \\
Random Search~\cite{fan_surrogate-assisted_2023} & 6.18$\pm$0.24          & 28.60$\pm$0.83          & 54.64$\pm$0.67          \\
REA~\cite{real_regularized_2019}                 & 5.94$\pm$0.29          & 27.28$\pm$0.72          & 54.01$\pm$0.51          \\
BANANAS~\cite{white_bananas_2021}                & 5.77$\pm$0.30          & 26.75$\pm$0.63          & 53.69$\pm$0.31          \\
GP bayesopt~\cite{white_bananas_2021}            & 5.84$\pm$0.31          & 26.95$\pm$0.75          & 53.75$\pm$0.34          \\
DNGO~\cite{fan_surrogate-assisted_2023}          & 5.92$\pm$0.26          & 27.34$\pm$0.67          & 54.00$\pm$0.48          \\
Bohamiann~\cite{white_bananas_2021}              & 5.91$\pm$0.26          & 27.35$\pm$0.66          & 53.97$\pm$0.48          \\
GCN Predictor~\cite{white_bananas_2021}          & 6.26$\pm$0.33          & 28.57$\pm$0.77          & 54.64$\pm$0.78          \\
BONAS~\cite{fan_surrogate-assisted_2023}         & 5.68$\pm$0.15          & 26.70$\pm$0.46          & 53.69$\pm$0.30          \\
NPENAS-SSRL~\cite{wei_self-supervised_2021}      & 5.68$\pm$0.19          & 26.53$\pm$0.30          & 54.17$\pm$0.60          \\
NPENAS-CCL~\cite{wei_self-supervised_2021}       & 5.68$\pm$0.19          & \textbf{26.51$\pm$0.23} & 54.39$\pm$0.41          \\
SAENAS-NE~\cite{fan_surrogate-assisted_2023}         & 5.66$\pm$0.12          & 26.54$\pm$0.20          & 53.64$\pm$0.26          \\
ReNAS~\cite{xu_renas_2021}                           & 6.01$\pm$0.25          & 27.88$\pm$0.79          & 54.03$\pm$0.49          \\
\hline
SiamNAS(ours)                                        & \textbf{5.63$\pm$0.00} & 27.39$\pm$0.31          & 54.07$\pm$0.08          \\ \hline
SiamNAS-transfer(ours)                               & \textbf{5.63$\pm$0.00} & 27.03$\pm$0.19          & 53.63$\pm$0.12          \\ \hline
Oracle                                               & 5.63                   & 26.49                   & 52.69                   \\ \hline
\end{tabular}
\end{scriptsize}
}
\end{table}

\begin{table}[]
\centering
\caption{Comparison of NAS algorithm runtime on NAS-Bench-201.}
\label{tab:runtime_201}
\resizebox{0.8\linewidth}{!}{%
\begin{tabular}{c|c|c}
\hline
\multirow{2}{*}{NAS Algorithm} & Time Cost    & \multirow{2}{*}{Method} \\ 
                         & (GPU Days) \\ \hline
DARTS~\cite{chu2021darts}                        & 4   & Gradient   \\
BANANAS~\cite{white_bananas_2021}                        & 2    & BO + neural predictor  \\
NAR~\cite{guo_generalized_2022}                         & 0.002  & Sampling + neural predictor    \\
ENAS~\cite{pham_efficient_2018}                         & 0.15 & Gradient     \\
SETN~\cite{9009811}                        & 0.35 & Gradient + neural predictor     \\
ReNAS~\cite{xu_renas_2021}                       & 0.001 & EA  + neural predictor      \\ \hline
SiamNAS(ours)                       & 0.01  & EA + neural predictor     \\ \hline
\end{tabular}
}
\end{table}





\section{Conclusion and Future Work}\label{sec:conc}

This article presents a novel surrogate model based on an ensemble of Siamese network blocks, each trained to predict the outcome of pairwise comparisons between candidate architectures. This approach is applied within the context of multi-objective NAS, aiming to identify accurate models with limited parameters and computational cost. Designed to be efficient and straightforward to train, the current surrogate model achieves an accuracy of approximately 92\% on predicting the outcome of pairwise comparisons. To demonstrate its potential, the model was integrated into the non-dominated sorting process of a multi-objective NAS framework, which seamlessly replaces the crowding distance calculation with a simple heuristic rule based on candidate architecture size. This integration allows SiamNAS to rely solely on the surrogate model, avoiding the need to train models during the search process.

The results from this proof-of-concept study on NAS-Bench-201, using three datasets, highlight the effectiveness of the proposed algorithm in achieving excellent performance while significantly reducing computational costs. As this work primarily serves as a foundation, future research will involve testing the proposed framework on real-world architectures beyond the constraints of benchmark datasets, further validating its applicability and robustness.

The Siamese network surrogate model introduced in this study holds promising potential for extending its application to multi-tasking optimisation \cite{10188456}, particularly in the context of generating Sets of Pareto Sets (SOS) \cite{10611967}. In SOS problems, where the objective is to create task-specific Pareto-optimal sets across heterogeneous environments, efficient methods for predicting dominance relationships are critical. A natural extension of this work could involve adapting the surrogate model to multi-tasking scenarios by leveraging its ability to learn and generalise dominance patterns across diverse tasks. This could facilitate the identification of commonalities between tasks, accelerate convergence to high-quality solutions, and enable effective knowledge transfer within evolutionary multitasking frameworks.

Moreover, future research could explore incorporating task-\\specific constraints or addressing dynamic optimisation problems where task environments evolve over time. By extending the capabilities of the surrogate model to handle these complexities, it could play a pivotal role in improving the efficiency and scalability of multi-tasking optimisation frameworks. Such advancements would not only enhance the practicality of the proposed model in real-world applications but also contribute to the broader development of multi-objective optimisation strategies.

\begin{acks}
    This project is supported by Ningbo Digital Port Technologies Key Lab, Ningbo Science and Technology Bureau (Project ID 2023Z237 and 2022Z217), the National Research Foundation, Singapore under its AI Singapore Programme (AISG Award No: AISG3-RP-2022-031) and the Jiangsu Distinguished Professor Programme.
\end{acks}

\bibliographystyle{ACM-Reference-Format}
\bibliography{sample-base}

\end{document}